\documentclass[runningheads]{llncs}

\usepackage[T1]{fontenc}
\usepackage{makecell}
\usepackage{float}
\usepackage{multirow}
\usepackage{hyperref}

\renewcommand{\figurename}{Fig.}
\usepackage{tabularx}
\usepackage{graphicx}
\usepackage{amsmath}
\usepackage{amssymb}
\usepackage{subcaption}

\usepackage[english]{babel}
\usepackage[subtle,bibnotes=tight]{savetrees}

\addto\extrasenglish{%
}

\usepackage{xargs}
\usepackage[colorinlistoftodos,prependcaption,textsize=tiny]{todonotes}
\newcommandx{\unsure}[2][1=]{\todo[linecolor=red,backgroundcolor=red!25,bordercolor=red,#1]{#2}}
\newcommandx{\change}[2][1=]{\todo[linecolor=blue,backgroundcolor=blue!25,bordercolor=blue,#1]{#2}}
\newcommandx{\info}[2][1=]{\todo[linecolor=OliveGreen,backgroundcolor=OliveGreen!25,bordercolor=OliveGreen,#1]{#2}}
\newcommandx{\improvement}[2][1=]{\todo[linecolor=Plum,backgroundcolor=Plum!25,bordercolor=Plum,#1]{#2}}
\newcommandx{\thiswillnotshow}[2][1=]{\todo[disable,#1]{#2}}
\newcommandx{\todoinline}[2][1=]{\todo[inline,#1]{#2}}

\begin{document}
\title{Neuro-Symbolic Process Anomaly Detection}
%
%
\author{Devashish Gaikwad\inst{1}\orcidID{0000-0001-6029-8863} \and Wil M. P. van der Aalst\inst{1}\orcidID{0000-0002-0955-6940} \and
Gyunam Park\inst{2}\orcidID{0000-0001-9394-6513}}
\authorrunning{D. Gaikwad  et al.}
%
\institute{RWTH Aachen University, Aachen, Germany\\
    \email{devashish.gaikwad@rwth-aachen.de, wvdaalst@pads.rwth-aachen.de}
    \and
    Eindhoven University of Technology, Eindhoven, The Netherlands\\
    \email{g.park@tue.nl}
}
\maketitle              
\begin{abstract}
    Process anomaly detection is an important application of process mining for identifying deviations from the normal behavior of a process.
Neural network-based methods have recently been applied to this task, learning directly from event logs without requiring a predefined process model.
However, since anomaly detection is a purely statistical task, these models fail to incorporate human domain knowledge.
As a result, rare but conformant traces are often misclassified as anomalies due to their low frequency, which limits the effectiveness of the detection process.
Recent developments in the field of neuro-symbolic AI have introduced Logic Tensor Networks (LTN) as a means to integrate symbolic knowledge into neural networks using real-valued logic.
In this work, we propose a neuro-symbolic approach that integrates domain knowledge into neural anomaly detection using LTN and Declare constraints.
Using autoencoder models as a foundation, we encode Declare constraints as soft logical guiderails within the learning process to distinguish between anomalous and rare but conformant behavior.
Evaluations on synthetic and real-world datasets demonstrate that our approach improves F1 scores even when as few as 10 conformant traces exist, and that the choice of Declare constraint and by extension human domain knowledge significantly influences performance gains.

\end{abstract}

\section{Introduction}

Process anomaly detection is a crucial application of process mining that helps organizations identify deviations from expected behavior in their processes \cite{bib:bezerra_anomaly_2009}.
It Not only helps uncover inefficiencies and bottlenecks, it also plays a vital role in ensuring compliance with regulatory standards.
An anomaly in a process is a deviation from its expected or defined behavior.
In this digital age, every activity in a process is recorded in the form of event logs. Therefore, process anomaly detection can be performed by analyzing them using process mining and machine learning techniques \cite{bib:nolle_unsupervised_2016_DAE,bib:nolle_binet_2022}.

If a pre-discovered process model exists, conformance checking can be used to assign fitness scores to traces (event sequences) to quantify their conformance \cite{bib:bezerra_anomaly_2009}.
However, in many cases, no process model is available, or the event log is very noisy and contains many deviations from the expected behavior.
In such cases, connectionist models (i.e., neural networks) can be used to model the expected behavior from the event log.
Current neural network based approaches rely on complex models such as LSTMs and Transformers to learn the control flow patterns from the event log and detect anomalies based on reconstruction errors \cite{bib:nolle_binet_2018,bib:nolle_binet_2022,BIB:johannes-lstm}.
These approaches also suffer from shortcomings such as requiring large amounts of event data, misclassification of rare but conformant traces, and lack the ability to incorporate human domain knowledge for interpretability \cite{bib:nolle_binet_2022}.

\begin{figure}[ht]
    \centering
    \includegraphics[width=\textwidth]{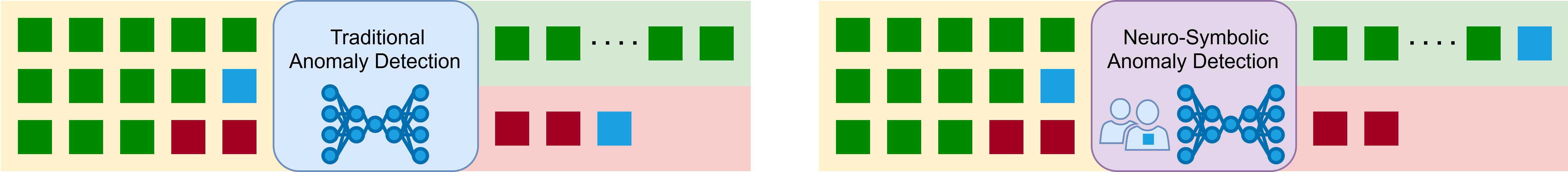}
    \caption{Comparison of classification of frequent and conformant traces (green tiles), rare but conformant traces (blue tile) and infrequent anomalous traces (red tiles).}
    \label{fig:introduction_diagram}
\end{figure}

Neuro-symbolic Artificial Intelligence (AI) is an emerging field that combines the strengths of both connectionist methods and symbolic methods (logic-based reasoning) and can aid in fine-tuning neural network based methods \cite{bib:neurosymbolic_ai}.
Event logs often exhibit noise and skewed frequency distributions, where common behaviors dominate and rare but conformant executions are underrepresented.
Since anomaly detection is inherently a statistical problem, the frequency of occurrence of traces plays a crucial role in determining whether a trace is conformant or not.
This results in misclassification of legitimate behavior as anomalous and increases data requirements.
As seen in \autoref{fig:introduction_diagram}, 
the frequently occurring behavior is classified as normal, while all other behavior is classified as anomalous even when it is compliant but rare. Neuro-symbolic AI techniques allow us to inject domain knowledge about rare but conformant traces into the neural network model, thus improving the anomaly detection performance.

In this paper, we propose a novel approach to enhance process anomaly detection using Logic Tensor Networks (LTN) \cite{bib:badreddine_logic_2022}, a neuro-symbolic AI framework, and Declare constraints, a declarative process modeling approach, allowing us to combine the strengths of both connectionist methods and symbolic human domain knowledge.
First, encode the control flow features of the event logs and train an autoencoder model on it to learn to reconstruct the traces.
Then, we mine the Declare constraints from the event log to capture the control flow patterns and select the most relevant ones using domain knowledge.
Crucially, we inject the domain knowledge into the autoencoder using LTN to optimize both the reconstruction error and the satisfaction of the Declare constraints for rare but conformant traces.
Finally, we use this autoencoder model with proven heuristic-based anomaly detection techniques \cite{bib:nolle_binet_2022} to detect anomalies in the event log.

The core idea is to leverage domain knowledge in the form of Declare constraints to fine-tune the autoencoder,
thus improving its ability to accurately reconstruct both common and rare yet conformant traces.
We evaluate our approach on both synthetic and real-world event logs that have been injected with anomalies, while using F1 scores as the evaluation metric for comparison with baseline models.
Our results show that our approach significantly improves the anomaly detection performance of autoencoders with LTN as compared to the baseline approach, starting with as few as 10 rare but conformant traces.
Furthermore, we demonstrate the effectiveness of different types of Declare constraints in improving the anomaly detection performance.

This paper is structured as follows: In \autoref{sec:related_works}, we give an overview of process anomaly detection and neuro-symbolic techniques and explore their use cases. 
Next, in \autoref{sec:main}, we explain the steps of our approach along with the relevant backgrounds them.
In \autoref{sec:evaluation}, we show the improvements and robustness of our approach. 
\autoref{sec:conclusion} concludes the paper.

\section{Related work}
\label{sec:related_works}

In this section, we introduce the background to process anomaly detection and neuro-symbolic AI techniques that can mitigate the shortcomings of current neural network based process anomaly detection methods.

\subsection{Process anomaly detection}

Process anomaly detection is a subfield of predictive process monitoring that focuses on identifying deviant behavior in normal process executions.
By analyzing event logs, process anomaly detection models are able to identify and flag unusual patterns that deviate from expected behavior.

As compared to newer neural network based approaches, classical methods in process mining for anomaly detection depend on a pre-discovered process model from the event log. These methods often focus on conformance checking. Most proposed methods such as Petri net discovery based anomaly detection \cite{bib:bezerra_anomaly_2009} work by
discovering the process model with the highest fitness score and then performing conformance checking against the model by classifying traces that do not fit the model as anomalous.


Nolle et al.~\cite{bib:nolle_unsupervised_2016_DAE} introduced denoising autoencoders and reconstruction error for anomaly detection in process mining. 
This work was extended by \cite{bib:nolle_binet_2018} which introduced BINet, a recurrent neural network based model that sequentially predicts the next event in a trace. It also introduced threshold heuristics for reconstruction error.
Finally, the same authors introduced a version with LSTM and attention \cite{bib:attention} mechanism in \cite{bib:nolle_binet_2022} which could generalize better.
Another LSTM based autoencoder model is proposed by \cite{BIB:johannes-lstm}, which predicts the next step in a sliding window of a trace and compares it with the actual next step.

A notable attempt to use Declare constraints in neural methods is seen in \cite{bib:de_smedt_predictive_2024}, where the authors introduce Processes-as-Movies. The processes are represented as a set of Declare constraints and the evolution of these Declare constraints is modeled using LSTM-based autoencoders.

In all of the above neural network based models, it can be seen that the architectures are based on the autoencoder-based trace reconstruction or next event prediction paradigm, while domain knowledge is generally not incorporated.

\subsection{Neuro-symbolic AI}
Neuro-symbolic AI combines neural network based learning with symbolic reasoning, creating systems that can robustly learn from data and reason using injected knowledge.
In the case of process anomaly detection, neuro-symbolic AI can help inject domain knowledge about the process into the neural network model, thus improving its ability to accurately reconstruct rare but conformant traces.

Garcez et al.~\cite{bib:neurosymbolic_ai} define a taxonomy of neuro-symbolic AI techniques based on the degree of integration between neural and symbolic components.
Categorized as a tightly-coupled hybrid system, DeepProblog combines a logic solver with neural networks by
using neural networks as predicates in a declarative logic program and then performing inference using the built-in solver.
By constraining the weights of the neurons in a recurrent neural network, Logical Neural Networks (LNN) \cite{bib:LNN} create a 1-to-1 correspondence between neurons and the elements of the logical formulas e.g., AND or OR gates or atoms. LNN is categorized as a compilation of symbolic knowledge into neural architecture according to the taxonomy of \cite{bib:neurosymbolic_ai}. There is a tight mapping between symbolic rules and the network structure.

Using fully differentiable logical operators, Logic Tensor Networks (LTN) \cite{bib:badreddine_logic_2022} allow the use of neural networks with outputs in the range $[0, 1]$ as predicates in a first-order logic formula.
Symbolic knowledge is directly incorporated into the neural network's loss function, thus acting as a soft constraint to the underlying neural architecture in the learning process for creating compliant models.
Garcez et al.~\cite{bib:neurosymbolic_ai} categorize it as a neuro-symbolic technique with embedded symbolic constraints. Although LTN requires symbolic formulas  to be manually constructed in advance for reasoning, it provides a flexible framework for integrating neural approaches and symbolic reasoning as a single system.


\section{Neuro-symbolic process anomaly detection}
\label{sec:main}

The core of our method is to combine neural models (autoencoders) and symbolic models (Declare constraints) using Logic Tensor Networks (LTN) to improve process anomaly detection.
Our approach addresses the limitation of purely statistical anomaly detection methods, i.e., their tendency to misclassify rare but conformant traces as anomalies due to low frequency in the event log.

\begin{figure}[]
    \centering
    \includegraphics[width=\textwidth]{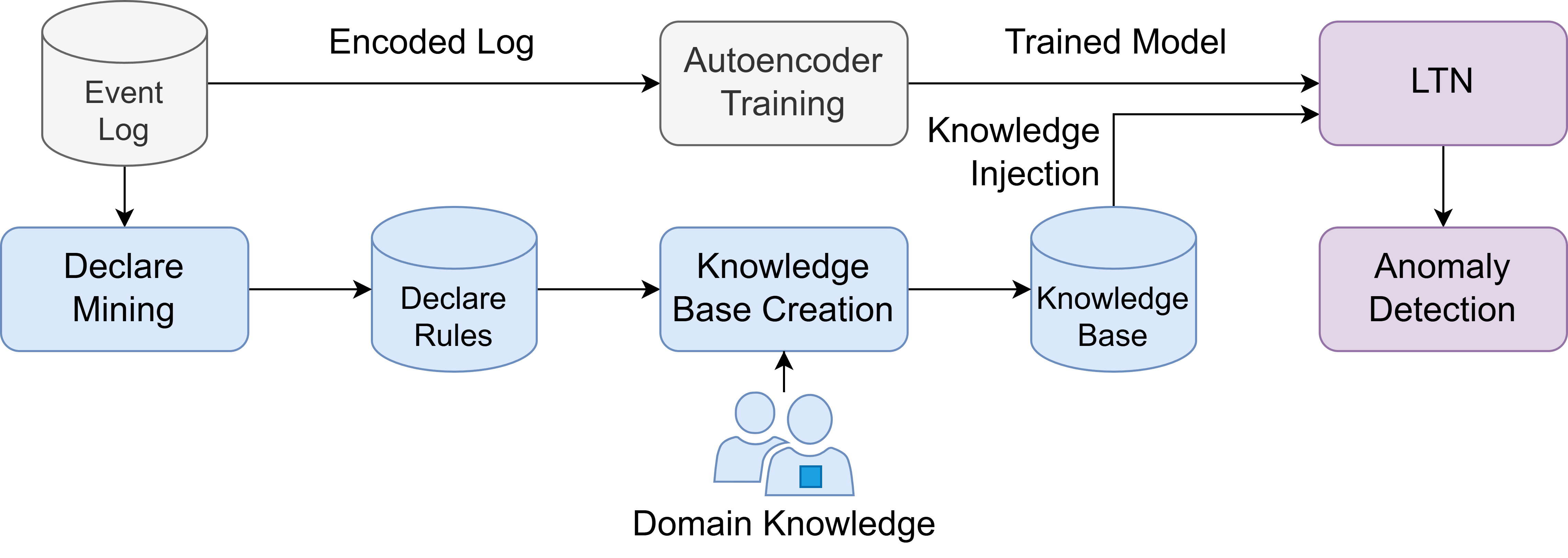}
    \caption{Overview of the proposed method. The event log is used both to pretrain an autoencoder for trace reconstruction and to mine Declare constraints. Domain knowledge guides the selection of constraints representing rare but conformant behavior, which are then injected into the autoencoder via LTN for fine-tuning. The enhanced model performs anomaly detection based on reconstruction error.}
    \label{fig:method_diagram}
\end{figure}

\autoref{fig:method_diagram} provides an overview of the proposed method. The process begins with the event log, which serves two purposes: first, it is encoded and used to pretrain an autoencoder that learns to reconstruct normal process traces; second, it is analyzed through Declare mining to extract declarative constraints that capture the process's control flow patterns. These mined constraints are filtered using domain knowledge to create a knowledge base representing rare but conformant behavior. This knowledge is then injected into the pretrained autoencoder using LTN, which fine-tunes the model to better reconstruct traces satisfying the selected Declare constraints. The resulting LTN-enhanced autoencoder performs anomaly detection based on reconstruction error.

\subsection{Autoencoder pretraining for trace reconstruction}
\label{sec:main_ae}
The foundation of our approach is a denoising autoencoder trained to reconstruct process traces. An autoencoder consists of an encoder that maps input traces to a lower-dimensional latent representation, and a decoder that reconstructs the original trace from this representation. The model is trained to minimize the reconstruction error between the input and output traces.

For trace encoding, we use one-hot encoding for the activity and resource attributes of each event. Since temporal ordering information is already preserved in the trace sequence, we do not encode event IDs or timestamps. This encoded representation serves as input to the autoencoder, which learns to reconstruct it through standard backpropagation.

The key principle of autoencoder-based anomaly detection is that the model learns to accurately reconstruct common patterns observed during training, while traces containing anomalous patterns result in higher reconstruction errors. After training, traces are classified as anomalous if their reconstruction error exceeds a predefined threshold. 

However, this purely statistical approach has a fundamental limitation: rare but conformant traces also produce high reconstruction errors and are thus misclassified as anomalies. To address this, we do not directly use the initially trained autoencoder for anomaly detection. Instead, we \emph{pretrain} it on the event log to establish a baseline understanding of normal trace reconstruction. In the subsequent steps, we enrich this pretrained model with domain knowledge to improve its ability to distinguish between true anomalies and rare conformant behavior.

\subsection{Knowledge base creation using Declare constraints}
\label{sec:main_declare_constraints}
To capture domain knowledge about valid process behavior, we leverage Declare constraints~\cite{bib:declare_full_support_4384001}. 
Declare constraints are formally specified through Linear Temporal Logic (LTL) formulas \cite{bib:model_checking}. 
Constraints representing ordering relations between activities are associated with two statistical measures: \emph{support}, which indicates how frequently the constraint's antecedent occurs in the event log, and \emph{confidence}, which measures the proportion of times the constraint is satisfied when its antecedent is activated.

We extract Declare constraints from the event log using established Declare mining algorithms \cite{bib:efficient_discovery_declare}. 
The mining process yields a comprehensive set of constraints along with their support and confidence values.
To construct our knowledge base, we filter these mined constraints based on three criteria: domain knowledge, support values, and confidence values.

Since our goal is to identify constraints representing rare but conformant behavior, we focus on constraints with \emph{low support and high confidence}. 
This combination indicates that while the constraint's antecedent rarely occurs (low support), it is almost always satisfied when activated (high confidence). 
We then apply domain knowledge to select the most relevant constraints from this filtered set, choosing those that best represent legitimate rare executions in the specific process context. 
These selected constraints constitute our knowledge base, which we subsequently integrate into the autoencoder training process.

\begin{figure}
    \centering
    \includegraphics[width=\textwidth]{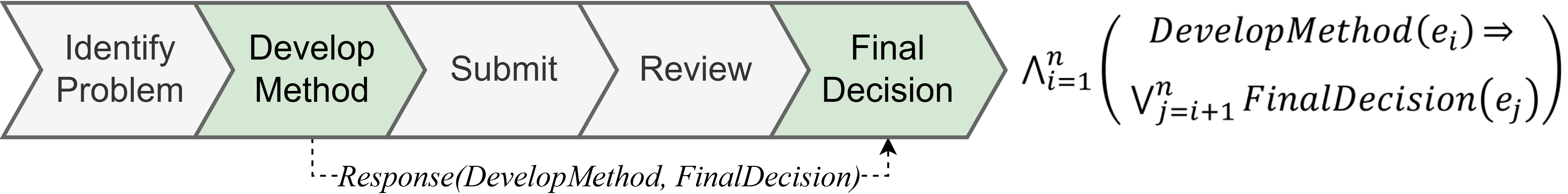}
    \caption{Example from the Paper event log showing the Response Declare constraint. If activity \textit{Develop Method} occurs, then \textit{Final Decision} must eventually follow.}
    \label{fig:declare_response_constraint}
\end{figure}

For example, as seen in \autoref{fig:declare_response_constraint}, the \textit{Response(Develop Method, Final Decision)} constraint states that if activity \textit{Develop Method} occurs in a trace, then activity \textit{Final Decision} must eventually follow it.
When represented in linear temporal logic, this constraint can be expressed as: $ \square (DevelopMethod \implies \lozenge FinalDecision) $,
which translates into first-order logic as $\wedge_{i=1}^n ( DevelopMethod\allowbreak (e_i) \implies \vee_{j=i+1}^n FinalDecision(e_j) )$ where $e_i$ and $e_j$ are events in the trace. We use the first-order logic representations of Declare constraints from \cite{bib:maggi_data-aware_2023}.

\subsection{Logic Tensor Networks for knowledge injection}
\label{sec:main_ltn}

\subsubsection{Background on real-valued logic}
Logic Tensor Networks (LTN) \cite{bib:badreddine_logic_2022} provide a neuro-symbolic framework that combines symbolic logic with neural network learning through fully differentiable logical operations. LTN grounds First-Order Logic (FOL) onto data using neural networks and fuzzy (real-valued) logic, enabling the integration of symbolic constraints into neural optimization.

As compared to symbolic FOL where logical values are binary, LTN uses a grounding function ($\mathcal{G}$) that maps symbols to their corresponding representations in a real-valued vector space, as well as fuzzy logic implementations of $\diamond \in \{\neg\} \text{ , } \circ \in \{\land, \lor, \implies, \iff \}$ to connect predicates, functions, logical formulas, and $\mathcal{Q} \in \{\forall, \exists\}$ to aggregate the values of each formula.

Specifically, $\{\exists, \forall\}$ are implemented using aggregation functions $A_{pM}$ (power-mean) and $A_{pME}$ (power-mean-error) respectively.

\begin{equation}
    \label{eq:power_mean}
    \tag{$\exists{x}$}
    A_{pM}\left(x_1, ..., x_n\right) = \left(\frac{1}{n}\sum_{i=1}^{n}x_i^p\right)^{\frac{1}{p}}
\end{equation}
\begin{equation}
    \label{eq:power_mean_error}
    \tag{$\forall{x}$}    
    A_{pME}\left(x_1, ..., x_n\right) = 1 - \left(\frac{1}{n}\sum_{i=1}^{n}(1-x_i)^p\right)^{\frac{1}{p}}
\end{equation}

Intuitively, $A_{pM}$ ($\exists$) is dominated by the maximum value (when $p$ is large), capturing that ``there exists at least one'' instance satisfying the predicate, i.e., a single high truth value makes the overall expression true. 
$A_{pME}$ ($\forall$) measures the aggregated error $(1-x_i)$ and is sensitive to violations, ensuring $\forall x: P(x)$ is satisfied only when all instances have high truth values, i.e., a single low value significantly reduces the overall satisfiability.

\begin{figure}[h]
    \centering
    \includegraphics[width=\textwidth]{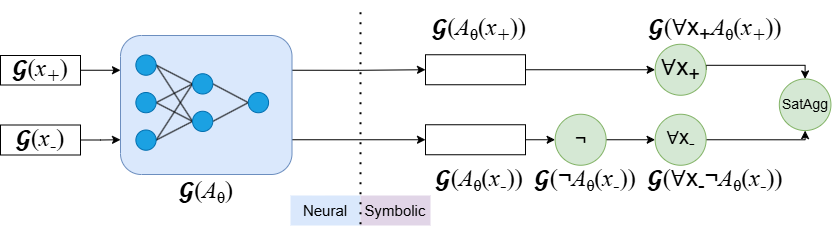}
    \caption{Learning a binary classifier predicate $A_{\theta}$ using LTN. The neural network outputs are interpreted as truth values and aggregated using logical quantifiers to compute overall satisfiability, which guides parameter learning.}
    \label{fig:ltn_prelim_learning}
\end{figure}

To illustrate LTN's learning mechanism, consider the binary classification example in \autoref{fig:ltn_prelim_learning}, where we learn a classifier $A_{\theta}$ that distinguishes positive points ($x_+$) from negative points ($x_-$) in 2D space. In the neural part, the predicate $A_{\theta}$ (implemented as a neural network with parameters $\theta$) evaluates each point and returns a truth value in $[0,1]$. For positive examples $x_+$, the output should be close to 1; for negative examples $x_-$, the output should be close to 0.

In the symbolic part, we express the learning objective using FOL: $\forall x_+ : A_{\theta}(x_+) \land \forall x_- : \neg A_{\theta}(x_-)$. The fuzzy aggregator $\forall x_+$ (implemented as $A_{pME}$) aggregates the truth values across all positive examples, ensuring all are classified as positive. Similarly, for negative examples, $\neg A_{\theta}(x_-) = 1 - A_{\theta}(x_-)$ should be close to 1, and $\forall x_-$ aggregates these negated truth values. The overall satisfiability is computed by aggregating both quantifier results using another $A_{pME}$ aggregator (denoted $SatAgg$), which quantifies how well the entire FOL formula is satisfied. The neural network parameters $\theta$ are learned by maximizing this overall satisfiability through gradient-based optimization, effectively using logical constraints as soft guidelines for learning.

\subsubsection{Integrating Declare constraints into autoencoder training}
We inject domain knowledge into the pretrained autoencoder using LTN by representing Declare constraints as FOL formulas and evaluating their satisfiability on reconstructed traces. 
The key insight is to interpret the autoencoder's output, i.e., a reconstructed trace, as a sequence of activity probability distributions that can serve as truth values for logical predicates.

Specifically, for each position $i$ in a reconstructed trace of length $n$, the autoencoder outputs a probability distribution over all possible activities. We define predicates $P_1, \ldots, P_n$ where $P_i(t, Activity)$ returns the truth value (probability) that a specific activity occurs at the $i^{th}$ position in trace $t$. These predicates bridge the neural and symbolic components: the autoencoder provides the raw probability values, while LTN uses them to evaluate Declare constraints expressed in FOL.

\begin{figure}[htb]
    \centering
    \includegraphics[width=\textwidth]{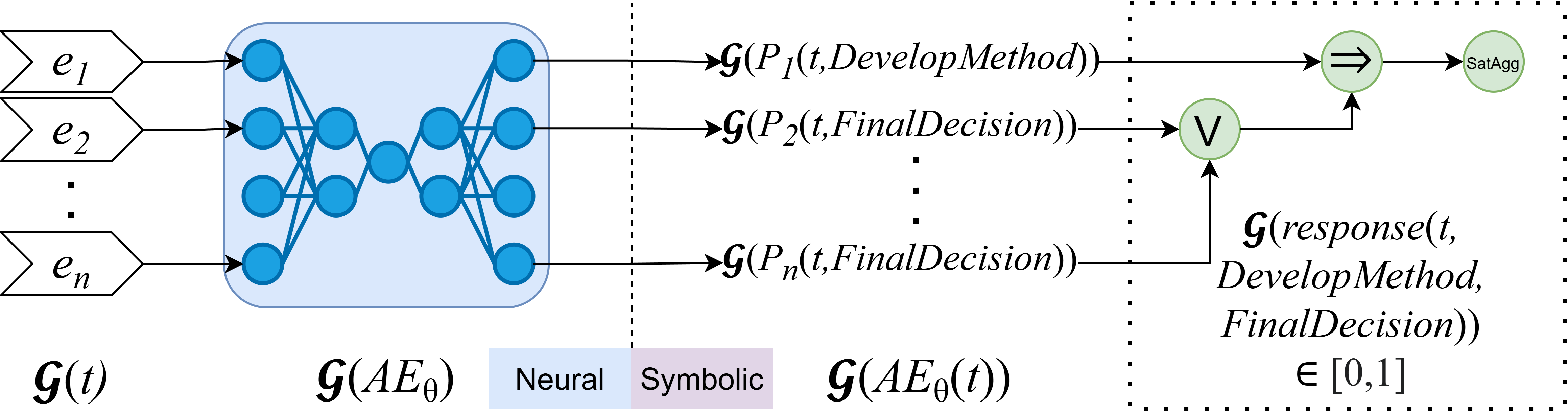}
    \caption{Fine-tuning the autoencoder $AE_{\theta}$ with LTN. Predicates $P_1$ to $P_n$ extract activity probabilities from reconstructed traces, which are used to evaluate Declare constraint satisfiability and guide learning.}
    \label{fig:method_ae_pred_ltn}
\end{figure}

\autoref{fig:method_ae_pred_ltn} illustrates the fine-tuning process using the \textit{Response(DevelopMethod, FinalDecision)} constraint as an example. The process operates in two complementary parts:
\begin{itemize}
    \item \textbf{Neural part:} The autoencoder $AE_{\theta}$ takes an encoded trace as input and produces a reconstructed trace. Each position in the reconstruction contains probability values for all possible activities, learned through the standard reconstruction objective.
    \item \textbf{Symbolic part:} We evaluate the Declare constraint's satisfiability on the reconstructed trace using its FOL representation. For the Response constraint, this means checking whether, for every position where \textit{Develop Method} has high probability, there exists a subsequent position with high probability for \textit{Final Decision}. This evaluation uses the predicates $P_i$ to extract relevant probabilities and applies LTN's fuzzy logic operators and aggregators to compute an overall satisfiability score in $[0,1]$.
\end{itemize}

To guide fine-tuning, we partition our training data into two sets based on the selected Declare constraint: $t_+$ contains traces that satisfy the constraint (rare but conformant traces we want to preserve), and $t_-$ contains traces that violate it (potential anomalies). The training objective becomes:
$$\mathcal{L}_{LTN} = \lambda_{rec} \cdot \mathcal{L}_{reconstruction} + \lambda_{sat} \cdot \left( \forall t_+ : \Phi(t_+) \land \forall t_- : \neg \Phi(t_-) \right)$$
, where $\Phi$ represents the Declare constraint in FOL, $\mathcal{L}_{reconstruction}$ is the standard reconstruction error, and $\lambda_{rec}, \lambda_{sat}$ are weighting hyperparameters. This objective encourages the autoencoder to maintain its reconstruction ability while learning to satisfy the Declare constraint on $t_+$ traces and violate it on $t_-$ traces, effectively teaching the model to recognize and properly reconstruct rare but valid patterns.

Through this fine-tuning process, the autoencoder learns to produce lower reconstruction errors for traces satisfying the domain knowledge constraints, even if they are statistically rare in the training data. Different Declare constraints can be used to inject their respective domain knowledge, with the choice of constraint significantly influencing the model's ability to distinguish between anomalous and rare but conformant behavior.

\subsection{Anomaly detection with the LTN-enhanced autoencoder}
After fine-tuning with LTN, we use the enhanced autoencoder for anomaly detection following the standard reconstruction error paradigm. For each trace in the evaluation set, we compute its reconstruction error as the mean squared difference between the encoded input and the autoencoder's output. 
Traces with reconstruction errors exceeding a predefined threshold are classified as anomalies.

The crucial difference from baseline autoencoder approaches is that our LTN-enhanced model has been explicitly trained to recognize and accurately reconstruct rare but conformant traces, i.e., those satisfying the injected Declare constraints. 
Consequently, such traces produce lower reconstruction errors despite their statistical rarity, reducing false positive classifications. 
Meanwhile, truly anomalous traces that violate process rules continue to produce high reconstruction errors, as they deviate from both statistical patterns and the symbolic constraints embedded in the model.

We follow the threshold-setting recommendations from \cite{bib:nolle_binet_2022}, which analyze the distribution of reconstruction errors across varying threshold values to determine optimal cutoffs that balance precision and recall. This heuristic approach ensures consistent evaluation across different event logs and baseline comparisons.


\section{Evaluation}
\label{sec:evaluation}

\subsection{Evaluation setup}

\subsubsection{Datasets}
We evaluated our approach on both synthetic and real-world event logs for anomaly detection. Details of the event logs used in the experiments are shown in \autoref{tab:datasets}.
We use the synthetic event logs from \cite{bib:nolle_binet_2022}, as they are the most widely used event logs for anomaly detection. In addition to these event logs, we also use real-world event logs from the Business Process Intelligence Challenge (BPIC):  BPIC12 \cite{bib:bpic12}, BPIC13 \cite{bib:bpic13}, and BPIC17 \cite{bib:bpic17}.
To focus on typical process behavior and ensure meaningful pattern learning, we preprocessed the event logs to only include traces under a specific trace length threshold, which is essential for both Declare constraint mining (to identify consistent patterns) and autoencoder training (to learn stable representations).
30\% of the cases in each event log were randomly selected and injected with anomalies using the anomaly injection framework from \cite{bib:nolle_binet_2022} to create anomalous traces for evaluation.

\begin{table}[]
    \centering
    \caption{Event logs used in evaluations}
    \label{tab:datasets}
    \begin{tabular}{lllllll}
        \hline
        Classification              & {Event Log}     & \makecell{Max. Trace                             \\Length} & \#Activities & \#Users & \#Cases & \#Events \\ \hline
        \multirow{8}{*}{Synthetic}  & Paper    & 17                   & 29  & 77  & 5000 & 60536  \\\cline{2-7}
                                    & P2P      & 16                   & 27  & 91  & 5000 & 53193  \\\cline{2-7}
                                    & Small    & 15                   & 41  & 89  & 5000 & 53437  \\\cline{2-7}
                                    & Medium   & 13                   & 65  & 105 & 5000 & 41991  \\\cline{2-7}
                                    & Large    & 17                   & 85  & 78  & 5000 & 67524  \\ \cline{2-7}
                                    & Huge     & 16                   & 109 & 93  & 5000 & 53210  \\ \cline{2-7}
                                    & Gigantic & 16                   & 155 & 111 & 5000 & 39829  \\ \cline{2-7}
                                    & Wide     & 12                   & 58  & 108 & 5000 & 41910  \\ \hline
        \multirow{3}{*}{Real-world} & BPIC12   & 14                   & 73  & -   & 6000 & 33598  \\ \cline{2-7}
                                    & BPIC13   & 14                   & 27  & -   & 1389 & 6124   \\ \cline{2-7}
                                    & BPIC17   & 24                   & 53  & -   & 6470 & 136584 \\ \hline
    \end{tabular}
\end{table}


\subsubsection{Setup}

In our evaluations \footnote{\url{https://github.com/DevashishX/LTNcoder}}, we compare two model approaches for anomaly detection.
Both approaches use the same autoencoder architecture and are trained exclusively on conformant traces (anomalies are reserved for evaluation only).
The key difference lies in how domain knowledge is incorporated:

\begin{itemize}
    \item \textbf{Baseline:} An autoencoder-only model trained on all traces in the event log without any domain knowledge integration,
    \item \textbf{Our approach:} A neuro-symbolic model using the same autoencoder architecture, pretrained on all the traces except rare but conformant traces, and then fine-tuned using LTN on rare but conformant traces to incorporate domain knowledge via Declare constraints.
\end{itemize}

Furthermore, to demonstrate the robustness of our approach, we create multiple versions of the same event log with varying number of rare but conformant traces. This allows us to evaluate the performance of the baseline and our approach in the same environments.

We use the F1 score as the primary evaluation metric for our anomaly detection models.
The F1 score is the harmonic mean of precision and recall, providing a balanced measure of a model's performance, especially in our scenario with imbalanced class distributions. 
Precision measures the accuracy of positive predictions. 
In our case, it is the proportion of detected anomalies that are actually correct. 
Recall measures the ability of a model to identify all relevant instances. Again in our case, recall is the proportion of actual anomalies that are correctly detected.

Our evaluation focuses on three key aspects:
\begin{itemize}
    \item \textbf{Synthetic event logs}: Evaluation of our approach on synthetic event logs, as these are conformant to a known process, 
    \item \textbf{Real-world event logs}: Evaluation of our approach on real-world event logs, these event logs already contain natural anomalies from real-world processes,
    \item \textbf{Effectiveness of Declare constraint selection}: Evaluation of approach against selection of different Declare constraints, as the effectiveness of domain knowledge is based on quality of selected constraints.
\end{itemize}

\subsection{Synthetic event logs}

\begin{table}[]
    \centering
    \caption{Evaluation of synthetic event logs}
    \label{tab:synthetic}
\begin{tabularx}{\textwidth}{ll*{8}{>{\centering\arraybackslash}X}}
    \hline
    \multirow{2}{*}{Event Log} 
        & \multirow{2}{*}{Method} 
        & \multicolumn{8}{c}{Anomaly detection F1 scores w.r.t. \# rare conformant traces} \\ \cline{3-10}
        &  
        & 10 & 25 & 50 & 100 & 150 & 200 & 250 & 300 \\ \hline
        \multirow{2}{*}{Paper} & Baseline & 0.38 & 0.38 & 0.39 & 0.38 & 0.4 & 0.42 & 0.39 & 0.43 \\ \cline{2-10}
        ~ & Our approach & \textbf{0.58} & \textbf{0.59} & \textbf{0.6} & \textbf{0.6} & \textbf{0.56} & \textbf{0.58} & \textbf{0.51} & \textbf{0.55} \\ \hline
        \multirow{2}{*}{P2P} & Baseline & 0.39 & 0.38 & 0.38 & 0.38 & 0.39 & 0.36 & 0.41 & 0.41 \\ \cline{2-10}
        ~ & Our approach & \textbf{0.46} & \textbf{0.48} & \textbf{0.51} & \textbf{0.49} & \textbf{0.46} & \textbf{0.48} & \textbf{0.46} & \textbf{0.44} \\ \hline
        \multirow{2}{*}{Small} & Baseline & 0.49 & 0.48 & 0.47 & 0.5 & 0.45 & 0.46 & 0.46 & 0.47 \\ \cline{2-10}
        ~ & Our approach & \textbf{0.59} & \textbf{0.58} & \textbf{0.58} & \textbf{0.57} & \textbf{0.56} & \textbf{0.56} & \textbf{0.54} & \textbf{0.5} \\ \hline
        \multirow{2}{*}{Medium} & Baseline & 0.25 & 0.24 & 0.25 & 0.23 & 0.23 & 0.25 & 0.25 & 0.28 \\ \cline{2-10}
        ~ & Our approach & \textbf{0.46} & \textbf{0.47} & \textbf{0.46} & \textbf{0.49} & \textbf{0.48} & \textbf{0.46} & \textbf{0.51} & \textbf{0.5} \\ \hline
        \multirow{2}{*}{Large} & Baseline & 0.27 & 0.26 & 0.32 & 0.31 & 0.32 & 0.34 & 0.35 & 0.36 \\ \cline{2-10}
        ~ & Our approach & \textbf{0.57} & \textbf{0.61} & \textbf{0.62} & \textbf{0.59} & \textbf{0.62} & \textbf{0.64} & \textbf{0.66} & \textbf{0.64} \\ \hline
        \multirow{2}{*}{Huge} & Baseline & 0.41 & 0.39 & 0.42 & 0.37 & 0.38 & 0.34 & 0.31 & 0.31 \\ \cline{2-10}
        ~ & Our approach & \textbf{0.46} & \textbf{0.56} & \textbf{0.54} & \textbf{0.52} & \textbf{0.5} & \textbf{0.52} & \textbf{0.57} & \textbf{0.54} \\ \hline
        \multirow{2}{*}{Gigantic} & Baseline & 0.36 & 0.33 & 0.34 & 0.34 & 0.33 & 0.37 & 0.38 & 0.33 \\ \cline{2-10}
        ~ & Our approach & \textbf{0.53} & \textbf{0.52} & \textbf{0.52} & \textbf{0.54} & \textbf{0.5} & \textbf{0.51} & \textbf{0.52} & \textbf{0.52} \\ \hline
        \multirow{2}{*}{Wide} & Baseline & \textbf{0.53} & \textbf{0.52} & \textbf{0.53} & \textbf{0.52} & \textbf{0.53} & \textbf{0.54} & \textbf{0.52} & \textbf{0.54} \\ \cline{2-10}
        ~ & Our approach & 0.52 & 0.5 & 0.5 & 0.43 & 0.45 & 0.45 & 0.52 & 0.47 \\ \hline
    \end{tabularx}
\end{table}

As seen in \autoref{tab:synthetic}, our proposed approach outperforms the baseline autoencoder model in 7 out of 8 synthetic event logs.
It is important to note that the absolute F1 scores are relatively low (ranging from 0.23 to 0.66), which reflects the inherent difficulty of anomaly detection in the commonly-used synthetic event logs, where anomalies are often subtle and the distinction between rare but conformant behavior and true anomalies is challenging.
However, the key finding is that our approach consistently outperforms the baseline across most event logs, demonstrating the effectiveness of incorporating domain knowledge through neuro-symbolic integration.
This improvement comes from a negligible decrease in precision and an increase in recall metrics, indicating that our approach better identifies true anomalies while maintaining similar precision.
Notably, this improvement is visible from the inclusion of as few as 10 rare but conformant traces.
The performance for the Wide event log shows no improvement. Since this event log is based on a wide process model with more trace variants than other event logs, autoencoders do not learn beyond their capacity limits. 
In fact, our approach results in decreased performance due to loss of generalization during fine-tuning.

\subsection{Real-world event logs}

\begin{table}[]
    \centering
    \caption{Evaluation of real-world event logs}
    \label{tab:real_world}
    \begin{tabularx}{\textwidth}{ll*{8}{>{\centering\arraybackslash}X}}
    \hline
        \multirow{2}{*}{Event Log} 
        & \multirow{2}{*}{Method} 
        & \multicolumn{8}{c}{Anomaly detection F1 scores w.r.t. \# rare conformant traces} \\ \cline{3-10}
        &  
        & 10 & 25 & 50 & 100 & 150 & 200 & 250 & 300 \\ \hline
        \multirow{2}{*}{BPIC12} & Baseline & 0.61 & 0.57 & 0.65 & 0.42 & 0.52 & 0.55 & 0.52 & 0.53 \\ \cline{2-10}
        ~ & Our approach & \textbf{0.66} & \textbf{0.75} & \textbf{0.78} & \textbf{0.53} & \textbf{0.62} & \textbf{0.61} & \textbf{0.62} & \textbf{0.63} \\ \hline
        \multirow{2}{*}{BPIC13} & Baseline & 0.27 & 0.29 & 0.26 & 0.28 & 0.29 & 0.29 & 0.28 & 0.3 \\ \cline{2-10}
        ~ & Our approach & \textbf{0.36} & \textbf{0.4} & \textbf{0.38} & \textbf{0.41} & \textbf{0.36} & \textbf{0.39} & \textbf{0.38} & \textbf{0.35} \\ \hline
        \multirow{2}{*}{BPIC17} & Baseline & \textbf{0.52} & 0.5 & 0.5 & 0.51 & \textbf{0.52} & 0.5 & 0.51 & 0.5 \\ \cline{2-10}
        ~ & Our approach & 0.51 & \textbf{0.52} & 0.5 & 0.51 & 0.51 & 0.5 & 0.51 & \textbf{0.51} \\ \hline
    \end{tabularx}
\end{table}

Evaluations in \autoref{tab:real_world} show that our proposed approach outperforms the baseline autoencoder model in 2 out of 3 real-world event logs.
Similar to synthetic event logs, the absolute F1 scores for real-world event logs are also relatively low (ranging from 0.26 to 0.78).
However, the key finding is that our approach consistently outperforms the baseline for BPIC12 and BPIC13, demonstrating the effectiveness of incorporating domain knowledge through neuro-symbolic integration in real-world settings.
This improvement again stems from increased recall, indicating that our approach better identifies true anomalies.
The improvement is again visible from the inclusion of as few as 10 rare but conformant traces.
Our approach does not show consistent improvements for the BPIC17 event log, where performance is comparable to the baseline with slight variations across different numbers of rare conformant traces. Similar to the Wide event log, this event log is too complex for the autoencoder to generalize effectively, limiting the benefits of domain knowledge integration.


\subsection{Ablation study: Impact of Declare constraint selection}

To understand the contribution of the different Declare constraints to the performance of our approach, we conduct an ablation study by systematically varying the Declare constraints used during LTN fine-tuning.
This allows us to analyze how the choice of domain knowledge (encoded as different Declare constraint templates) affects anomaly detection performance.
We evaluate seven different Declare constraints, in isolation, on the Paper and BPIC12 event logs, comparing the best performing baseline model against the best performing LTN model using our approach for each constraint. 

\subsubsection{Paper synthetic event log}

\begin{figure}[h]
    \centering
    \includegraphics[width=\textwidth]{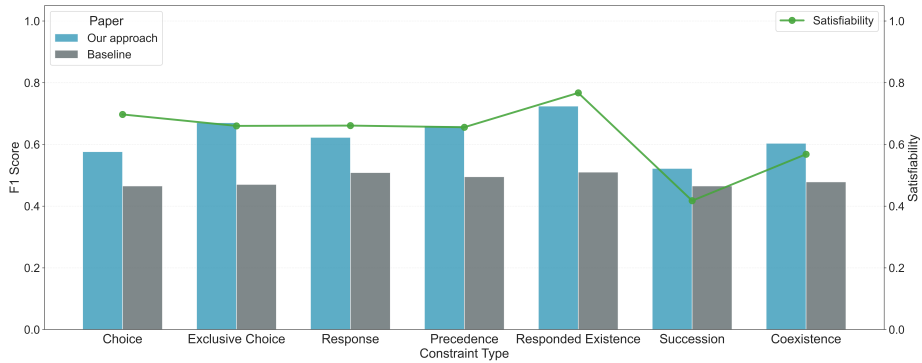}
    \caption{Impact of Declare constraints for the best performing models for Paper event log.}
    \label{fig:eval_declare_impact_paper}
\end{figure}

The ablation results in \autoref{fig:eval_declare_impact_paper} reveal that the performance improvement of our approach is highly dependent on the specific Declare constraint used.
The F1 score improvements vary significantly across different constraint types, demonstrating that not all domain knowledge is equally effective for anomaly detection.
Furthermore, the satisfiability of the Declare constraint after training also varies across constraints, indicating that some constraints are easier for the model to learn and satisfy than others.
This variation highlights the importance of careful constraint selection based on domain knowledge and the characteristics of the event log.

We observe several patterns in the results:
For constraints \textit{choice} and \textit{exclusive choice}, which do not have confidence values associated with them, we selected these constraints based on the known likelihood graph of the process model.
The \textit{succession} constraint shows low satisfiability after training, which can be attributed to its low confidence in the original event log, making it difficult for the model to satisfy during fine-tuning.
Despite this low satisfiability, the F1 score still improves slightly, suggesting that even partial constraint satisfaction can provide benefits through the averaging behavior of the autoencoder.

\subsubsection{BPIC12 real-world event log}

\begin{figure}[h]
    \centering
    \includegraphics[width=\textwidth]{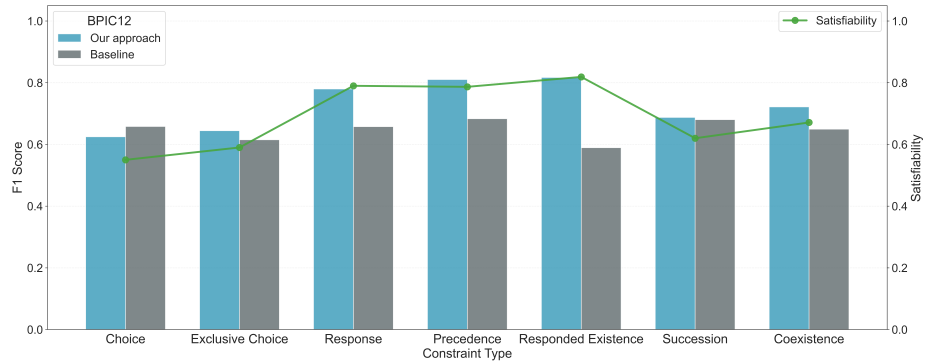}
    \caption{Impact of Declare constraints for the best performing models for BPIC12 event log.}
    \label{fig:eval_declare_impact_bpic12}
\end{figure}

We present the comparison between the best performing autoencoder-only model and the best performing model using our approach for BPIC12 event log and seven Declare Constraints in \autoref{fig:eval_declare_impact_bpic12}.  

The ablation results for BPIC12 in \autoref{fig:eval_declare_impact_bpic12} confirm the findings from the Paper event log: the performance improvement is again dependent on the specific Declare constraint used.
This consistency across both synthetic and real-world event logs demonstrates that the effectiveness of different Declare constraints is a fundamental characteristic of our approach, rather than being specific to a particular event log.
The variation in performance across constraints underscores the importance of domain knowledge in selecting appropriate Declare constraints that align with the process characteristics and the types of rare, but conformant, behavior we aim to preserve.

\section{Conclusion}
\label{sec:conclusion}

In this paper, we introduced a neuro-symbolic approach to anomaly detection in process mining, integrating symbolic reasoning with neural networks to incorporate domain knowledge directly into the learning process.
By leveraging Logic Tensor Networks and Declare constraints, we use domain knowledge in autoencoder-based anomaly detection.
Our framework allows for the integration of domain knowledge into model training rather than relying solely on feature engineering.
Through experiments on both synthetic and real-world event logs, we demonstrate that incorporating domain knowledge improves the detection of rare but conformant traces, even with as few as 10 examples and mitigates issues arising from skewed event logs.
Since the increase in F1 score also depends on the chosen Declare constraint, careful selection of relevant domain knowledge is crucial for optimal performance, highlighting the importance of human expertise.

In future work, we plan to increase the performance by using newer underlying models and by making the training process more efficient.
The dependency of LTN performance on the underlying neural network's generalization ability suggests the need for more complex architectures, such as graph-based or LSTM models. Since the injection of domain knowledge is fully contained in the LTN layer, any neural network which can produce a trace reconstruction can be improved using LTN.
Moreover, since real-valued logic does not allow for computational shortcuts like short-circuiting boolean expressions, investigating more efficient real-logic evaluation strategies and optimized constraint expressions will reduce the cost of computation during training.





%
%
%
\bibliographystyle{splncs04}
\bibliography{references}
\end{document}